\documentclass[11pt]{article}
\usepackage[a4paper,margin=2.5cm]{geometry}
\usepackage[T1]{fontenc}
\usepackage[utf8]{inputenc}
\usepackage{lmodern}
\usepackage{microtype}
\usepackage{amsmath,amssymb}
\usepackage{booktabs}
\usepackage{graphicx}
\usepackage{xcolor}
\usepackage[numbers,sort&compress]{natbib}
\usepackage[colorlinks=true,linkcolor=blue!50!black,citecolor=blue!50!black,urlcolor=blue!50!black]{hyperref}

\newcommand{\N}[1]{\ifcsname fsnum@#1\endcsname\csname fsnum@#1\endcsname\else\textbf{\textcolor{red}{??#1}}\fi}
\expandafter\def\csname fsnum@abLargeDice\endcsname{0.886}
\expandafter\def\csname fsnum@abLargeSec\endcsname{1.50}
\expandafter\def\csname fsnum@abSmallDice\endcsname{0.885}
\expandafter\def\csname fsnum@abSmallSec\endcsname{2.86}
\expandafter\def\csname fsnum@clinOneMax\endcsname{9.3}
\expandafter\def\csname fsnum@clinOneNet\endcsname{1.1}
\expandafter\def\csname fsnum@clinOneSec\endcsname{5.0}
\expandafter\def\csname fsnum@clinOneVRAM\endcsname{3.8}
\expandafter\def\csname fsnum@clinScans\endcsname{10}
\expandafter\def\csname fsnum@clinTwoMax\endcsname{4.3}
\expandafter\def\csname fsnum@clinTwoNet\endcsname{1.1}
\expandafter\def\csname fsnum@clinTwoSec\endcsname{2.5}
\expandafter\def\csname fsnum@clinTwoVRAM\endcsname{3.3}
\expandafter\def\csname fsnum@eqDiff\endcsname{+0.0001}
\expandafter\def\csname fsnum@eqHi\endcsname{+0.0004}
\expandafter\def\csname fsnum@eqLo\endcsname{-0.0002}
\expandafter\def\csname fsnum@eqP\endcsname{3.5\times10^{-5}}
\expandafter\def\csname fsnum@eqScans\endcsname{30}
\expandafter\def\csname fsnum@filtFalseRemoved\endcsname{26}
\expandafter\def\csname fsnum@filtMm\endcsname{337.5}
\expandafter\def\csname fsnum@filtRealRemoved\endcsname{12}
\expandafter\def\csname fsnum@filtSec\endcsname{0.10}
\expandafter\def\csname fsnum@filtSizes\endcsname{0, 10, 25, 50, 100, 200, 400, 800, 1600}
\expandafter\def\csname fsnum@filtValDiceOff\endcsname{0.878}
\expandafter\def\csname fsnum@filtValDiceOn\endcsname{0.882}
\expandafter\def\csname fsnum@filtVox\endcsname{100}
\expandafter\def\csname fsnum@fpFilter\endcsname{9}
\expandafter\def\csname fsnum@fpNoFilter\endcsname{35}
\expandafter\def\csname fsnum@fsBeatsSmallOrgans\endcsname{12}
\expandafter\def\csname fsnum@fsDice\endcsname{0.901}
\expandafter\def\csname fsnum@fsGapDefault\endcsname{0.008}
\expandafter\def\csname fsnum@fsLccDice\endcsname{0.900}
\expandafter\def\csname fsnum@fsLccNSD\endcsname{0.914}
\expandafter\def\csname fsnum@fsLccNetSec\endcsname{1.1}
\expandafter\def\csname fsnum@fsLccScans\endcsname{30}
\expandafter\def\csname fsnum@fsLccSec\endcsname{1.59}
\expandafter\def\csname fsnum@fsLccVRAM\endcsname{4.1}
\expandafter\def\csname fsnum@fsNSD\endcsname{0.915}
\expandafter\def\csname fsnum@fsNetSec\endcsname{1.1}
\expandafter\def\csname fsnum@fsParamsM\endcsname{18.0}
\expandafter\def\csname fsnum@fsScans\endcsname{30}
\expandafter\def\csname fsnum@fsSec\endcsname{1.52}
\expandafter\def\csname fsnum@fsVRAM\endcsname{4.1}
\expandafter\def\csname fsnum@fsVzeroDice\endcsname{0.886}
\expandafter\def\csname fsnum@fsVzeroNSD\endcsname{0.893}
\expandafter\def\csname fsnum@fsVzeroNetSec\endcsname{1.1}
\expandafter\def\csname fsnum@fsVzeroScans\endcsname{30}
\expandafter\def\csname fsnum@fsVzeroSec\endcsname{1.50}
\expandafter\def\csname fsnum@fsVzeroVRAM\endcsname{4.1}
\expandafter\def\csname fsnum@fsWeightsMB\endcsname{72}
\expandafter\def\csname fsnum@gpuAbdDice\endcsname{0.909}
\expandafter\def\csname fsnum@gpuAbdNSD\endcsname{0.927}
\expandafter\def\csname fsnum@gpuAbdScans\endcsname{30}
\expandafter\def\csname fsnum@gpuAbdSec\endcsname{3.38}
\expandafter\def\csname fsnum@gpuAbdVRAM\endcsname{9.4}
\expandafter\def\csname fsnum@gpuAllDice\endcsname{0.909}
\expandafter\def\csname fsnum@gpuAllNSD\endcsname{0.927}
\expandafter\def\csname fsnum@gpuAllScans\endcsname{30}
\expandafter\def\csname fsnum@gpuAllSec\endcsname{7.98}
\expandafter\def\csname fsnum@gpuAllVRAM\endcsname{10.5}
\expandafter\def\csname fsnum@gpuHAbdA\endcsname{4.5}
\expandafter\def\csname fsnum@gpuHAbdAIters\endcsname{57k}
\expandafter\def\csname fsnum@gpuHAbdB\endcsname{11.9}
\expandafter\def\csname fsnum@gpuHAbdBIters\endcsname{150k}
\expandafter\def\csname fsnum@gpuHAbdC\endcsname{20.4}
\expandafter\def\csname fsnum@gpuHAbdCIters\endcsname{256k}
\expandafter\def\csname fsnum@gpuHAbdTotal\endcsname{37}
\expandafter\def\csname fsnum@gpuHSpine\endcsname{15.3}
\expandafter\def\csname fsnum@gpuHSpineIters\endcsname{178k}
\expandafter\def\csname fsnum@gpuHThorax\endcsname{19.5}
\expandafter\def\csname fsnum@gpuHThoraxIters\endcsname{262k}
\expandafter\def\csname fsnum@gpuSmallDice\endcsname{0.884}
\expandafter\def\csname fsnum@gpuSmallNSD\endcsname{0.895}
\expandafter\def\csname fsnum@gpuSmallScans\endcsname{30}
\expandafter\def\csname fsnum@gpuSmallSec\endcsname{1.50}
\expandafter\def\csname fsnum@gpuSmallVRAM\endcsname{9.6}
\expandafter\def\csname fsnum@hFsBeatsTsNewFilterOrgans\endcsname{5}
\expandafter\def\csname fsnum@hFsBeatsTsNewOrgans\endcsname{6}
\expandafter\def\csname fsnum@hFsDice\endcsname{0.884}
\expandafter\def\csname fsnum@hFsGainFilter\endcsname{0.015}
\expandafter\def\csname fsnum@hFsGainOverOld\endcsname{0.017}
\expandafter\def\csname fsnum@hFsGapGpuAbd\endcsname{-0.008}
\expandafter\def\csname fsnum@hFsGapTsNew\endcsname{-0.003}
\expandafter\def\csname fsnum@hFsGapTsNewFilter\endcsname{0.003}
\expandafter\def\csname fsnum@hFsLeadGpuAbd\endcsname{0.008}
\expandafter\def\csname fsnum@hFsLeadTsNew\endcsname{0.003}
\expandafter\def\csname fsnum@hFsNSD\endcsname{0.900}
\expandafter\def\csname fsnum@hFsNSDthree\endcsname{0.936}
\expandafter\def\csname fsnum@hFsNoFilterDice\endcsname{0.869}
\expandafter\def\csname fsnum@hFsNoFilterNSD\endcsname{0.886}
\expandafter\def\csname fsnum@hFsNoFilterScans\endcsname{109}
\expandafter\def\csname fsnum@hFsNoFilterSec\endcsname{1.06}
\expandafter\def\csname fsnum@hFsNoFilterVRAM\endcsname{4.4}
\expandafter\def\csname fsnum@hFsOldDice\endcsname{0.867}
\expandafter\def\csname fsnum@hFsOldFilterDice\endcsname{0.876}
\expandafter\def\csname fsnum@hFsOldFilterNSD\endcsname{0.890}
\expandafter\def\csname fsnum@hFsOldFilterScans\endcsname{109}
\expandafter\def\csname fsnum@hFsOldFilterSec\endcsname{1.06}
\expandafter\def\csname fsnum@hFsOldFilterVRAM\endcsname{4.4}
\expandafter\def\csname fsnum@hFsOldNSD\endcsname{0.884}
\expandafter\def\csname fsnum@hFsOldScans\endcsname{109}
\expandafter\def\csname fsnum@hFsOldSec\endcsname{1.06}
\expandafter\def\csname fsnum@hFsOldVRAM\endcsname{4.4}
\expandafter\def\csname fsnum@hFsScans\endcsname{109}
\expandafter\def\csname fsnum@hFsSec\endcsname{1.16}
\expandafter\def\csname fsnum@hFsVRAM\endcsname{4.2}
\expandafter\def\csname fsnum@hGpuAbdDice\endcsname{0.876}
\expandafter\def\csname fsnum@hGpuAbdNSD\endcsname{0.895}
\expandafter\def\csname fsnum@hGpuAbdNSDthree\endcsname{0.930}
\expandafter\def\csname fsnum@hGpuAbdScans\endcsname{109}
\expandafter\def\csname fsnum@hGpuAbdSec\endcsname{2.25}
\expandafter\def\csname fsnum@hGpuAbdVRAM\endcsname{9.5}
\expandafter\def\csname fsnum@hTsNewDice\endcsname{0.881}
\expandafter\def\csname fsnum@hTsNewFilterDice\endcsname{0.887}
\expandafter\def\csname fsnum@hTsNewFilterNSD\endcsname{0.905}
\expandafter\def\csname fsnum@hTsNewFilterNSDthree\endcsname{0.936}
\expandafter\def\csname fsnum@hTsNewFilterScans\endcsname{109}
\expandafter\def\csname fsnum@hTsNewFilterSec\endcsname{4.30}
\expandafter\def\csname fsnum@hTsNewFilterVRAM\endcsname{5.9}
\expandafter\def\csname fsnum@hTsNewFolderDice\endcsname{0.881}
\expandafter\def\csname fsnum@hTsNewFolderNSD\endcsname{0.902}
\expandafter\def\csname fsnum@hTsNewFolderScans\endcsname{109}
\expandafter\def\csname fsnum@hTsNewFolderSec\endcsname{1.76}
\expandafter\def\csname fsnum@hTsNewFolderVRAM\endcsname{5.9}
\expandafter\def\csname fsnum@hTsNewNSD\endcsname{0.902}
\expandafter\def\csname fsnum@hTsNewNSDthree\endcsname{0.935}
\expandafter\def\csname fsnum@hTsNewScans\endcsname{109}
\expandafter\def\csname fsnum@hTsNewSec\endcsname{4.30}
\expandafter\def\csname fsnum@hTsNewVRAM\endcsname{5.9}
\expandafter\def\csname fsnum@roiAbdScans\endcsname{10}
\expandafter\def\csname fsnum@roiAbdSec\endcsname{28.9}
\expandafter\def\csname fsnum@roiSpineScans\endcsname{10}
\expandafter\def\csname fsnum@roiSpineSec\endcsname{18.3}
\expandafter\def\csname fsnum@spFsBeatsGpuOrgans\endcsname{19}
\expandafter\def\csname fsnum@spFsBeatsTsNewOrgans\endcsname{12}
\expandafter\def\csname fsnum@spFsDice\endcsname{0.894}
\expandafter\def\csname fsnum@spFsGapTsNew\endcsname{0.004}
\expandafter\def\csname fsnum@spFsLccDice\endcsname{0.888}
\expandafter\def\csname fsnum@spFsLccNSD\endcsname{0.926}
\expandafter\def\csname fsnum@spFsLccScans\endcsname{109}
\expandafter\def\csname fsnum@spFsLccSec\endcsname{1.22}
\expandafter\def\csname fsnum@spFsLccVRAM\endcsname{6.2}
\expandafter\def\csname fsnum@spFsNSD\endcsname{0.934}
\expandafter\def\csname fsnum@spFsNSDthree\endcsname{0.951}
\expandafter\def\csname fsnum@spFsScans\endcsname{109}
\expandafter\def\csname fsnum@spFsSec\endcsname{1.20}
\expandafter\def\csname fsnum@spFsVRAM\endcsname{6.2}
\expandafter\def\csname fsnum@spGpuDice\endcsname{0.888}
\expandafter\def\csname fsnum@spGpuNSD\endcsname{0.922}
\expandafter\def\csname fsnum@spGpuNSDthree\endcsname{0.939}
\expandafter\def\csname fsnum@spGpuScans\endcsname{109}
\expandafter\def\csname fsnum@spGpuSec\endcsname{1.28}
\expandafter\def\csname fsnum@spGpuVRAM\endcsname{10.0}
\expandafter\def\csname fsnum@spRare\endcsname{L6}
\expandafter\def\csname fsnum@spStructs\endcsname{25}
\expandafter\def\csname fsnum@spTsNewDice\endcsname{0.898}
\expandafter\def\csname fsnum@spTsNewNSD\endcsname{0.937}
\expandafter\def\csname fsnum@spTsNewNSDthree\endcsname{0.952}
\expandafter\def\csname fsnum@spTsNewScans\endcsname{109}
\expandafter\def\csname fsnum@spTsNewSec\endcsname{4.30}
\expandafter\def\csname fsnum@spTsNewVRAM\endcsname{5.9}
\expandafter\def\csname fsnum@thFsBeatsTsNewOrgans\endcsname{3}
\expandafter\def\csname fsnum@thFsDice\endcsname{0.917}
\expandafter\def\csname fsnum@thFsGapGpu\endcsname{0.005}
\expandafter\def\csname fsnum@thFsGapTsNew\endcsname{0.004}
\expandafter\def\csname fsnum@thFsNSD\endcsname{0.903}
\expandafter\def\csname fsnum@thFsScans\endcsname{109}
\expandafter\def\csname fsnum@thFsSec\endcsname{1.10}
\expandafter\def\csname fsnum@thFsVRAM\endcsname{3.2}
\expandafter\def\csname fsnum@thGpuDice\endcsname{0.922}
\expandafter\def\csname fsnum@thGpuNSD\endcsname{0.912}
\expandafter\def\csname fsnum@thGpuScans\endcsname{109}
\expandafter\def\csname fsnum@thGpuSec\endcsname{2.25}
\expandafter\def\csname fsnum@thGpuVRAM\endcsname{9.5}
\expandafter\def\csname fsnum@thStructs\endcsname{7}
\expandafter\def\csname fsnum@thTsNewDice\endcsname{0.921}
\expandafter\def\csname fsnum@thTsNewNSD\endcsname{0.913}
\expandafter\def\csname fsnum@thTsNewScans\endcsname{109}
\expandafter\def\csname fsnum@thTsNewSec\endcsname{4.30}
\expandafter\def\csname fsnum@thTsNewVRAM\endcsname{5.9}
\expandafter\def\csname fsnum@tsDefaultCpuPct\endcsname{78}
\expandafter\def\csname fsnum@tsDefaultDice\endcsname{0.909}
\expandafter\def\csname fsnum@tsDefaultNSD\endcsname{0.927}
\expandafter\def\csname fsnum@tsDefaultNetPct\endcsname{13}
\expandafter\def\csname fsnum@tsDefaultNetSec\endcsname{9.7}
\expandafter\def\csname fsnum@tsDefaultScans\endcsname{30}
\expandafter\def\csname fsnum@tsDefaultSec\endcsname{74.8}
\expandafter\def\csname fsnum@tsDefaultVRAM\endcsname{9.6}
\expandafter\def\csname fsnum@tsFastCpuPct\endcsname{67}
\expandafter\def\csname fsnum@tsFastDice\endcsname{0.816}
\expandafter\def\csname fsnum@tsFastNSD\endcsname{0.828}
\expandafter\def\csname fsnum@tsFastNetPct\endcsname{5}
\expandafter\def\csname fsnum@tsFastNetSec\endcsname{0.8}
\expandafter\def\csname fsnum@tsFastScans\endcsname{30}
\expandafter\def\csname fsnum@tsFastSec\endcsname{14.3}
\expandafter\def\csname fsnum@tsFastVRAM\endcsname{5.4}
\expandafter\def\csname fsnum@tsFastestCpuPct\endcsname{76}
\expandafter\def\csname fsnum@tsFastestDice\endcsname{0.702}
\expandafter\def\csname fsnum@tsFastestNSD\endcsname{0.605}
\expandafter\def\csname fsnum@tsFastestNetPct\endcsname{3}
\expandafter\def\csname fsnum@tsFastestNetSec\endcsname{0.3}
\expandafter\def\csname fsnum@tsFastestScans\endcsname{30}
\expandafter\def\csname fsnum@tsFastestSec\endcsname{9.27}
\expandafter\def\csname fsnum@tsFastestVRAM\endcsname{0.8}
\expandafter\def\csname fsnum@tsNetParamsM\endcsname{31.2}
\expandafter\def\csname fsnum@tsSmallCpuPct\endcsname{80}
\expandafter\def\csname fsnum@tsSmallDice\endcsname{0.884}
\expandafter\def\csname fsnum@tsSmallNSD\endcsname{0.895}
\expandafter\def\csname fsnum@tsSmallNetParamsM\endcsname{13.3}
\expandafter\def\csname fsnum@tsSmallNetPct\endcsname{13}
\expandafter\def\csname fsnum@tsSmallNetSec\endcsname{9.1}
\expandafter\def\csname fsnum@tsSmallScans\endcsname{30}
\expandafter\def\csname fsnum@tsSmallSec\endcsname{72.3}
\expandafter\def\csname fsnum@tsSmallVRAM\endcsname{9.8}
\expandafter\def\csname fsnum@xGpuAbdOverFs\endcsname{2.2}
\expandafter\def\csname fsnum@xHGpuAbdOverHFs\endcsname{1.9}
\expandafter\def\csname fsnum@xHTsNewFolderOverHFs\endcsname{1.5}
\expandafter\def\csname fsnum@xHTsNewFolderOverSpFs\endcsname{1.5}
\expandafter\def\csname fsnum@xHTsNewFolderOverThFs\endcsname{1.6}
\expandafter\def\csname fsnum@xHTsNewOverHFs\endcsname{3.7}
\expandafter\def\csname fsnum@xHTsNewOverSpFs\endcsname{3.6}
\expandafter\def\csname fsnum@xHTsNewOverThFs\endcsname{3.9}
\expandafter\def\csname fsnum@xRoiAbdOverHFs\endcsname{25}
\expandafter\def\csname fsnum@xRoiSpineOverSpFs\endcsname{15}
\expandafter\def\csname fsnum@xSpGpuOverSpFs\endcsname{1.1}
\expandafter\def\csname fsnum@xThGpuOverThFs\endcsname{2.0}
\expandafter\def\csname fsnum@xTsDefaultOverFs\endcsname{49}
\expandafter\def\csname fsnum@xTsDefaultOverGpuAbd\endcsname{22}
\expandafter\def\csname fsnum@xTsDefaultOverGpuAll\endcsname{9.4}
\expandafter\def\csname fsnum@xTsDefaultOverGpuSmall\endcsname{50}
\expandafter\def\csname fsnum@xTsFastOverFs\endcsname{9.4}
\expandafter\def\csname fsnum@xTsSmallOverFs\endcsname{48}

\newif\ifdraft\draftfalse

\newcommand{\fs}{FennecSeg}
\newcommand{\ts}{TotalSegmentator}

\title{\fs: Compact CT Segmentation Models that Match \ts{} at a Fraction of the Compute}
\author{Mohammed El Amine Mokhtari \qquad Nour Islam Mokhtari \\[6pt]
{\normalsize PYCAD} \\[2pt]
{\normalsize\texttt{mohammed@pycad.co} \quad \texttt{nourislam@pycad.co}}}
\date{}

\begin{document}
\maketitle

\begin{abstract}
We present \fs, a family of compact 3D segmentation models for CT and an open library to run them.
Each \fs{} model covers one group of anatomical structures with a small residual-encoder U-Net (\N{fsParamsM}M parameters, \N{fsWeightsMB}\,MB), is trained on the public \ts{} dataset on a single consumer GPU (\N{gpuHSpine}--\N{gpuHThorax}\,GPU hours for a new group warm-started from an existing model), and runs entirely on the GPU at inference.
On the \N{hFsScans} scans of the \ts{} test split and an NVIDIA RTX 4070, the 14-structure abdominal model reaches a Dice score of \N{hFsDice} in \N{hFsSec}\,s per scan with at most \N{hFsVRAM}\,GB of GPU memory, including a small-piece filter chosen on validation scans, the spine model (vertebrae and sacrum) reaches \N{spFsDice} in \N{spFsSec}\,s, and the thorax model (five lung lobes, trachea and heart) \N{thFsDice} in \N{thFsSec}\,s.
We compare them with \ts{} v3.1, which segments 117 structures with one network and nnU-Net's new streaming inference (\N{hTsNewFolderSec}--\N{hTsNewSec}\,s per scan; Dice \N{hTsNewDice} on the abdomen, \N{spTsNewDice} on the spine and \N{thTsNewDice} on the thorax), and with \ts's earlier networks run in the same GPU pipeline as \fs, which separates the contribution of the model from that of the pipeline.
\fs{} is within 0.005 Dice of \ts{} v3.1 on all three groups (on the abdomen, \N{hFsLeadTsNew} above v3.1 out of the box and \N{hFsGapTsNewFilter} below it with the same filter), at \N{xHTsNewFolderOverSpFs} to \N{xHTsNewOverThFs} times lower time per scan for the structures it covers, and with less GPU memory for the abdominal and thorax models.
The library reads NIfTI and DICOM at any resolution and orientation, needs neither \ts{} nor nnU-Net, and loads versioned models by name.
Code and weights (Apache-2.0): \url{https://github.com/pycadd/FennecSeg} and \url{https://huggingface.co/amine0110/fennecseg}.
\end{abstract}

\section{Introduction}
\label{sec:intro}

Automatic segmentation of anatomical structures in CT is a routine building block for organ volumetry, opportunistic screening, radiotherapy and surgical planning, and the curation of large imaging datasets.
\ts{}~\citep{wasserthal2023totalsegmentator}, built on nnU-Net~\citep{isensee2021nnunet}, made this broadly accessible with open models for more than a hundred structures.

Its cost per scan has long been a practical limit: in our benchmark, the released version (2.18) takes \N{tsDefaultSec}\,s per CT on a consumer GPU.
As several groups have now shown, most of this time is spent around the networks rather than in them, in CPU pre-processing and in the export of the predictions~\citep{totalsegmentator_v3,nnunet_issue3079}.
Pipelines that keep inference on the GPU make the same weights an order of magnitude faster~\citep{fastsegmentator2026,yamagishi2026flashsegmentator,boussot2025totalsegkonfai}, and the forthcoming \ts{} v3.1 combines nnU-Net's new streaming inference with a single network for all 117 structures~\citep{isensee2026totalsegpr648,isensee2026streaming}.
Our own profiling, on a 12\,GB consumer GPU, agrees: the networks take \N{tsDefaultNetPct}\% of the runtime of \ts{} 2.18.

Once inference runs on the GPU, the remaining cost is the networks themselves.
\fs{} addresses that part: instead of one large model for all structures, it uses one small model per group of structures, so that a user pays only for the structures they need, and a new group can be trained cheaply.

\paragraph{This report describes:}
\begin{enumerate}
  \item \textbf{The \fs{} models}: a compact residual-encoder U-Net (\N{fsParamsM}M parameters) and a time-budgeted training recipe that fits a consumer GPU (Section~\ref{sec:method}).
  \item \textbf{A benchmark on a consumer GPU} (Sections~\ref{sec:setup}--\ref{sec:results}) against \ts{} as users run it (version 2.18 and v3.1) and against \ts's default and small networks inside the same GPU pipeline as \fs, which separates what the model contributes from what the pipeline contributes.
  On all \N{hFsScans} test scans, \fs{} reaches Dice \N{hFsDice} for 14 abdominal structures (\N{hFsSec}\,s per scan), \N{spFsDice} for the spine (\N{spFsSec}\,s) and \N{thFsDice} for the thorax (\N{thFsSec}\,s), against \N{hTsNewDice}, \N{spTsNewDice} and \N{thTsNewDice} for \ts{} v3.1 (\N{hTsNewFolderSec}--\N{hTsNewSec}\,s per scan for all 117 structures).
  \item \textbf{An open library} with versioned model packages that runs on any CT without \ts{} or nnU-Net installed (Section~\ref{sec:library}).
\end{enumerate}

\section{Related work}
\label{sec:related}

\paragraph{\ts{} and nnU-Net.}
\ts~\citep{wasserthal2023totalsegmentator} segments 117 CT structures with five nnU-Net~\citep{isensee2021nnunet} networks at 1.5\,mm, plus faster 3\,mm and 6\,mm modes; an MRI counterpart exists~\citep{dantonoli2025totalsegmentatormri}.
Version 3 of its dataset (1,939 scans, CC BY 4.0) is the training source of \fs.
The v3 development branch adds smaller residual-encoder networks (\texttt{model\_size="small"}), whose documentation notes that on a GPU the runtime is dominated by pre- and post-processing, and reports 105\,s for a large CT on an RTX 3090 before the latest changes~\citep{totalsegmentator_v3}.
nnU-Net's residual-encoder presets~\citep{isensee2024nnunetrevisited} and the \texttt{dynamic\_network\_architectures} library~\citep{dynamic_network_architectures} provide the building blocks we also use.

\paragraph{Faster inference for the same models.}
A public nnU-Net issue measured 8\,s of network time in a 305\,s prediction with a \ts{} model, most of the rest being the export of class probabilities on one CPU core~\citep{nnunet_issue3079}.
FastSegmentator~\citep{fastsegmentator2026} runs resampling, sliding-window inference and post-processing of \ts{} on the GPU, up to 9.6$\times$ faster with identical output; FlashSegmentator~\citep{yamagishi2026flashsegmentator} adds batching across scans and reports under one second per CT on a data-centre GPU at equal Dice; TotalSegmentator-KonfAI~\citep{boussot2025totalsegkonfai} reports 1.8--3.9$\times$.
nnU-Net's streaming inference fuses sliding window, resampling and arg-max into one GPU pass, and \ts{} v3.1 uses it with a single network for all 117 structures, reported at 10--14\,s per CT on an RTX 3090~\citep{isensee2026totalsegpr648,isensee2026streaming,totalsegmentator_v3}.
Compression by tensor decomposition~\citep{weber2024tucker} and post-training quantisation~\citep{qu2025ptq} reduce the cost of the networks themselves.
\fs{} uses GPU-resident inference of the same kind; its contribution is the models.

\paragraph{Efficient abdominal segmentation.}
The FLARE challenges rank abdominal organ segmentation by accuracy, runtime and GPU memory~\citep{ma2022flare21,ma2024flare22,ma2024flare23,flare2024proceedings}.
The FLARE~2023 winner segments 13 abdominal organs in 8.6\,s with 3.6\,GB of GPU memory at Dice 0.92 on a 2019 workstation GPU, using a low-resolution localisation step and a crop~\citep{ma2024flare23}; the 2024 edition added a CPU-only laptop track.
Cropping to a region and large spacings are standard in these methods, and we use the former.

\paragraph{Larger and general models.}
Transformer and large convolutional models (UNETR, Swin UNETR, nnFormer, MedNeXt, STU-Net)~\citep{hatamizadeh2022unetr,hatamizadeh2022swinunetr,zhou2021nnformer,roy2023mednext,huang2023stunet}, pre-trained or promptable models (SuPreM, the CLIP-driven universal model, SAT, VISTA3D)~\citep{li2024suprem,liu2024universal,zhao2023sat,he2025vista3d} and the MONAI whole-body bundle~\citep{monai_wholebody_ct,cardoso2022monai} aim at accuracy or generality rather than cost per scan; most are larger than \fs{} by one to two orders of magnitude.
A hierarchical sparse-sampling method reaches about 2\,s per CT on a CPU, at substantially lower Dice~\citep{guo2025sparse}.

\section{Method}
\label{sec:method}

\subsection{Model}
\label{sec:model}
\fs{} uses the residual-encoder U-Net of the nnU-Net ecosystem~\citep{isensee2024nnunetrevisited} with a smaller configuration chosen for speed:
five resolution stages with 24, 48, 96, 192 and 320 feature maps (\ts's default networks: six stages, 32 to 320),
1, 2, 2, 2 and 2 residual blocks in the encoder,
one convolution per decoder stage instead of two,
instance normalisation and leaky ReLU, and no deep supervision.
The model has \N{fsParamsM}M parameters (\N{fsWeightsMB}\,MB in 32-bit floats), against \N{tsNetParamsM}M for each of the five networks of \ts{} 2.18, two of which are needed for our 14 abdominal structures, \N{tsSmallNetParamsM}M for each of its small networks, and 102M for the single network of \ts{} v3.1~\citep{isensee2026totalsegpr648}.
One model covers one group of structures (a \emph{task}); the label set, input spacing, intensity statistics and default inference settings are stored with the weights, so that new groups need no code changes.

\subsection{Inference}
\label{sec:pipeline}
Inference stays on the GPU from the moment the CT is read until the label map is written, as in recent fast \ts{} pipelines~\citep{fastsegmentator2026,yamagishi2026flashsegmentator,isensee2026streaming}:
the volume is reoriented to RAS and, if its spacing differs from 1.5\,mm by more than 1\%, resampled with trilinear interpolation on the GPU;
CT values are clipped to the 0.5 and 99.5 percentiles of the training foreground and standardised;
windows are combined with Gaussian weights in half precision;
and only the 8-bit label map is copied back to the CPU.
When the input was resampled, the class scores are resampled back to the input grid on the GPU, slab by slab to bound memory, so that labels are produced at the input resolution rather than by nearest-neighbour upsampling of a coarse label map.
Two further choices are specific to \fs.
It crops the volume to the bounding box of the body (voxels above $-500$\,HU, with a margin of 8 voxels) before the sliding window, and it uses larger windows with less overlap than \ts: $160^3$ voxels with step 0.8, while it is trained on $128^3$ patches.
On the first \N{fsScans} test scans, the first abdominal model reached Dice \N{abSmallDice} with $128^3$ windows and step 0.5 (\N{abSmallSec}\,s per scan) and \N{abLargeDice} with $160^3$ windows and step 0.8 (\N{abLargeSec}\,s), so the larger, less overlapping windows halve the time at no cost in accuracy.

\subsection{Training}
\label{sec:training}
Models are trained from the public \ts{} dataset (Section~\ref{sec:data}) on $128^3$ patches at 1.5\,mm with batch size 2, mixed precision, and SGD with Nesterov momentum 0.99, weight decay $3\cdot10^{-5}$ and an initial learning rate of 0.01 with polynomial decay (power 0.9).
The loss is the sum of cross-entropy and a batch soft Dice loss averaged over the foreground classes present in the batch.
Averaging over all classes instead rewards predicting nothing for the classes a patch does not contain, since an empty prediction of an absent class scores a Dice of 1: with 26 vertebrae, of which a $128^3$ patch shows only a few, that variant collapsed to predicting background, and for abdomen14 the present-class version improved the earlier release (Section~\ref{sec:results}).
Half of the patches are centred near a voxel of a structure present in the scan, chosen uniformly over the present structures so that small structures are seen often; the other half are placed at random.
Data augmentation runs on the GPU: random rotations up to $\pm15^\circ$ and isotropic scaling between 0.8 and 1.2 (probability 0.3), intensity scaling, shifts and Gaussian noise.
There is no mirroring, because left and right structures have different labels.

The learning-rate schedule follows the wall clock rather than an iteration count: training runs until a given end time, and the polynomial decay is applied to the fraction of that time already used.
This makes the schedule independent of the hardware's speed, which is hard to estimate in advance when data loading and validation share the machine.
The released abdomen14 model (0.3.0) was trained in three runs: \N{gpuHAbdAIters} iterations from scratch (\N{gpuHAbdA}\,h); \N{gpuHAbdBIters} iterations (\N{gpuHAbdB}\,h) warm-started from the first, with the weakest structures (gallbladder, pancreas, adrenal glands, esophagus, duodenum, portal and splenic veins) sampled twice as often (release 0.2.0, Dice loss over all classes); and \N{gpuHAbdCIters} iterations (\N{gpuHAbdC}\,h) warm-started from 0.2.0 with the present-class Dice loss, \N{gpuHAbdTotal}\,GPU hours in all.
spine26 was trained in one run of \N{gpuHSpineIters} iterations (\N{gpuHSpine}\,h) warm-started from abdomen14 0.2.0, and thorax7 in one run of \N{gpuHThoraxIters} iterations (\N{gpuHThorax}\,h) warm-started from abdomen14 0.3.0 with the present-class Dice loss.
New tasks can be warm-started from a model of another task: all weights except the final classification layer are copied.

\subsection{Post-processing}
\label{sec:post}
Small isolated false detections, mostly of structures that are absent from the scan (for example a few voxels of liver in a scan that ends above it), cost a full Dice point each for that scan and structure.
\fs{} therefore removes, for every structure, connected pieces (26-connectivity) smaller than a fixed volume.
The volume was chosen on the 20 validation scans, never on test scans: among \N{filtSizes} voxels at 1.5\,mm, \N{filtVox} voxels (\N{filtMm}\,mm$^3$) gave the highest validation Dice (\N{filtValDiceOn}, against \N{filtValDiceOff} without the filter).
The library applies it on the output grid with the threshold converted from mm$^3$, so it does not depend on the input resolution; it takes \N{filtSec}\,s per scan on the CPU and can be switched off.
Keeping only the largest piece of each organ, which we also tried, did not help.

\section{Experimental setup}
\label{sec:setup}

\subsection{Data}
\label{sec:data}
We use version 3.0.0 of the public \ts{} dataset~\citep{wasserthal2023totalsegmentator} (1,939 CT scans with reference masks for 117 structures, CC BY 4.0) and its official split.
Models are trained on the training split minus 20 scans held out to monitor training; all results are on the official test split, which no model saw.
The \ts{} 2.18 command, which takes over a minute per scan, was run on the first \N{fsScans} test scans in all its modes and on the first \N{roiAbdScans} with \texttt{roi\_subset}.
All scans are stored at 1.5\,mm.
Clinical-resolution inputs are covered in Section~\ref{sec:library}.

\paragraph{Structures.} \emph{abdomen14}: spleen, right and left kidney, gallbladder, liver, stomach, pancreas, right and left adrenal gland, esophagus, duodenum, aorta, inferior vena cava, and the portal and splenic veins.
\emph{spine26}: the 25 vertebrae from C1 to L6 and the sacrum, the same structures as \ts's vertebrae network. \N{spRare} appears in only one test scan, so the spine means are over the other \N{spStructs} structures.
\emph{thorax7}: the five lung lobes, the trachea and the heart.

\subsection{Compared methods}
\begin{itemize}
  \item \textbf{\ts{} 2.18} (latest PyPI release) with the v3 weights, as users run it: default (five networks at 1.5\,mm), fast (3\,mm), small networks (\texttt{model\_size="small"}), and the default mode restricted to our structures with \texttt{roi\_subset}.
  \item \textbf{\ts{} v3.1} (development branch \texttt{totalseg\_v3}, commit \texttt{a5329b9}, with nnU-Net 2.9 streaming inference, commit \texttt{9cea112}): one network for 117 structures, called once per scan through its Python interface, which loads the network each time as the command line does, and once on a folder of all test scans with the network loaded once (time divided by the number of scans). Its labels are mapped to the \ts{} 2.18 label ids by name.
  \item \textbf{\ts{} networks in our pipeline}: \ts's default and small networks (v3 weights, unchanged) run with the same GPU-resident inference as \fs{} (Section~\ref{sec:pipeline}) and \ts's own window overlap (step 0.8), loading only the networks that contain the evaluated structures. Comparing these rows with \fs{} isolates the effect of the model.
  \item \textbf{\fs} (this work).
\end{itemize}

\subsection{Metrics and timing}
Accuracy is measured with the Dice coefficient and the normalised surface Dice at 2\,mm tolerance~\citep{nikolov2021clinically}, which we also report at 3\,mm, the tolerance used by \ts~\citep{totalsegmentator_v3}, per scan and structure; a structure absent from both the reference and the prediction is skipped.
We report the mean over structures of the per-structure mean over scans.
Times are end to end per scan, from reading the input file to writing the label map, averaged over scans after one warm-up scan, and include model loading only where the method reloads models for every scan (the \ts{} command).
Peak GPU memory is the maximum memory allocated by PyTorch.
Hardware: one NVIDIA RTX 4070 (12\,GB) in a desktop PC with an Intel Core i7-14700K and 64\,GB of RAM, Windows 11, Python 3.12, PyTorch 2.11 with CUDA 12.8; training and all timings use this machine.

\section{Results}
\label{sec:results}

\begin{table}[t]
\centering
\caption{The 14 abdominal structures on all \N{hFsScans} test scans (RTX 4070). Time per scan is end to end. \ts{} v3.1 segments all 117 structures; ``per scan'' calls it once per scan, as the command line does, and ``folder'' runs it once on all scans with the network loaded once. ``Pipeline'' runs \ts's unchanged v3 weights in our GPU pipeline. ``+ filter'': the size filter of Section~\ref{sec:post} applied to the output (about \N{filtSec}\,s per scan, not included in the v3.1 time). $^\dagger$\ts{} 2.18 command restricted to these structures with \texttt{roi\_subset}, timed on the first \N{roiAbdScans} test scans.}
\label{tab:main}
\begin{tabular}{lrrrr}
\toprule
Method & s/scan & Dice & NSD & GPU GB \\
\midrule
\ts{} 2.18 \texttt{roi\_subset} (command)$^\dagger$ & \N{roiAbdSec} & -- & -- & \\
\ts{} v3.1, one network (per scan) & \N{hTsNewSec} & \N{hTsNewDice} & \N{hTsNewNSD} & \N{hTsNewVRAM} \\
\ts{} v3.1, one network (folder) & \N{hTsNewFolderSec} & \N{hTsNewDice} & \N{hTsNewNSD} & \N{hTsNewVRAM} \\
\ts{} v3.1 + filter & & \textbf{\N{hTsNewFilterDice}} & \textbf{\N{hTsNewFilterNSD}} & \\
\ts{} v3, 2 networks (pipeline) & \N{hGpuAbdSec} & \N{hGpuAbdDice} & \N{hGpuAbdNSD} & \N{hGpuAbdVRAM} \\
\midrule
\fs{} abdomen14 0.2.0 & \N{hFsOldSec} & \N{hFsOldDice} & \N{hFsOldNSD} & \N{hFsOldVRAM} \\
\fs{} abdomen14 0.3.0, no filter & \N{hFsNoFilterSec} & \N{hFsNoFilterDice} & \N{hFsNoFilterNSD} & \N{hFsNoFilterVRAM} \\
\fs{} abdomen14 0.3.0 (released, library) & \textbf{\N{hFsSec}} & \N{hFsDice} & \N{hFsNSD} & \textbf{\N{hFsVRAM}} \\
\bottomrule
\end{tabular}
\end{table}

\begin{table}[t]
\centering
\caption{The spine on all \N{spFsScans} test scans: mean over the \N{spStructs} structures present in at least five test scans (vertebrae C1--L5 and sacrum; \N{spRare} appears in one scan). Rows as in Table~\ref{tab:main}; $^\dagger$timed on the first \N{roiSpineScans} test scans.}
\label{tab:spine}
\begin{tabular}{lrrrr}
\toprule
Method & s/scan & Dice & NSD & GPU GB \\
\midrule
\ts{} 2.18 \texttt{roi\_subset} (command)$^\dagger$ & \N{roiSpineSec} & -- & -- & \\
\ts{} v3.1, one network (per scan) & \N{hTsNewSec} & \textbf{\N{spTsNewDice}} & \textbf{\N{spTsNewNSD}} & \N{hTsNewVRAM} \\
\ts{} v3.1, one network (folder) & \N{hTsNewFolderSec} & \textbf{\N{spTsNewDice}} & \textbf{\N{spTsNewNSD}} & \N{hTsNewVRAM} \\
\ts{} v3 vertebrae network (pipeline) & \N{spGpuSec} & \N{spGpuDice} & \N{spGpuNSD} & \N{spGpuVRAM} \\
\midrule
\fs{} spine26 & \textbf{\N{spFsSec}} & \N{spFsDice} & \N{spFsNSD} & \N{spFsVRAM} \\
\bottomrule
\end{tabular}
\end{table}

\begin{table}[t]
\centering
\caption{The thorax (five lung lobes, trachea, heart) on all \N{thFsScans} test scans. Rows as in Table~\ref{tab:main}; the \ts{} v3 row runs its organ and cardiac networks in our pipeline (the same two networks and run as the pipeline row of Table~\ref{tab:main}).}
\label{tab:thorax}
\begin{tabular}{lrrrr}
\toprule
Method & s/scan & Dice & NSD & GPU GB \\
\midrule
\ts{} v3.1, one network (per scan) & \N{hTsNewSec} & \N{thTsNewDice} & \textbf{\N{thTsNewNSD}} & \N{hTsNewVRAM} \\
\ts{} v3.1, one network (folder) & \N{hTsNewFolderSec} & \N{thTsNewDice} & \textbf{\N{thTsNewNSD}} & \N{hTsNewVRAM} \\
\ts{} v3 organ and cardiac networks (pipeline) & \N{thGpuSec} & \textbf{\N{thGpuDice}} & \N{thGpuNSD} & \N{thGpuVRAM} \\
\midrule
\fs{} thorax7 0.1.0 (released, library) & \textbf{\N{thFsSec}} & \N{thFsDice} & \N{thFsNSD} & \textbf{\N{thFsVRAM}} \\
\bottomrule
\end{tabular}
\end{table}

\begin{figure}[t]
\centering
\includegraphics[width=\linewidth]{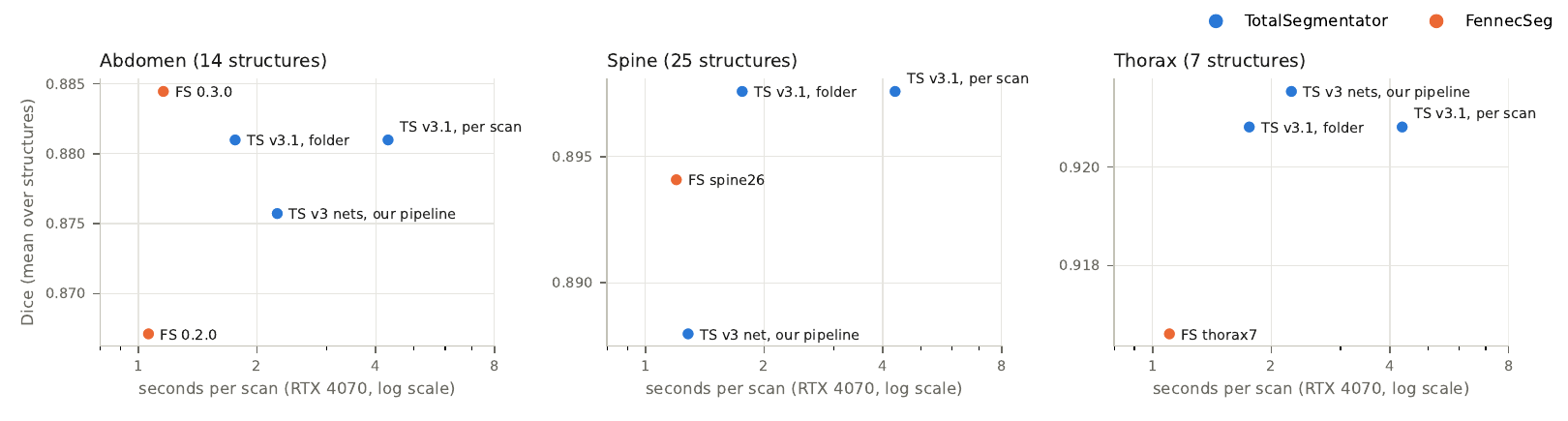}
\caption{Time per scan against Dice on all \N{hFsScans} test scans (RTX 4070). FS: \fs; TS: \ts. The \ts{} 2.18 command (\N{tsDefaultSec}\,s per scan, or \N{roiAbdSec}\,s with \texttt{roi\_subset}) is outside the plotted range.}
\label{fig:speed}
\end{figure}

\paragraph{Abdomen.}
The released abdomen14 model (0.3.0), run with the \texttt{fennecseg} library, takes \N{hFsSec}\,s per scan, \N{xHTsNewOverHFs}$\times$ less than \ts{} v3.1 called per scan and \N{xHTsNewFolderOverHFs}$\times$ less than v3.1 on a folder, with \N{hFsVRAM}\,GB of GPU memory against \N{hTsNewVRAM}\,GB.
Its Dice (\N{hFsDice}) is \N{hFsLeadTsNew} above \ts{} v3.1 out of the box and \N{hFsLeadGpuAbd} above \ts's v3 networks in our pipeline.
Applying the same size filter to \ts{} v3.1 raises it to \N{hTsNewFilterDice}, \N{hFsGapTsNewFilter} above \fs; against that, \fs{} is better or equal on \N{hFsBeatsTsNewFilterOrgans} of the 14 structures and about 0.01 behind on the liver, gallbladder, right adrenal gland and right kidney (Table~\ref{tab:organs}).
The two changes since 0.2.0 add \N{hFsGainOverOld} Dice: the present-class loss improves the structures present in a scan, notably the adrenal glands, spleen and esophagus, but also produced more small false detections of absent structures (\N{fpNoFilter} scan--structure pairs against \N{fpFilter} after filtering); the size filter adds \N{hFsGainFilter}.
At \ts's 3\,mm tolerance the normalised surface Dice is \N{hFsNSDthree} for \fs, \N{hTsNewNSDthree} for \ts{} v3.1 and \N{hTsNewFilterNSDthree} for v3.1 with the filter (spine: \N{spFsNSDthree} against \N{spTsNewNSDthree}).
The filter removed \N{filtFalseRemoved} false detections and \N{filtRealRemoved} true structures, all smaller than its threshold: small adrenal glands and organs that only just enter the scan.

\paragraph{Spine.}
\fs{} spine26 reaches Dice \N{spFsDice}, \N{spFsGapTsNew} below \ts{} v3.1 and above \ts's v3 vertebrae network (\N{spGpuDice}), which it beats on \N{spFsBeatsGpuOrgans} of the \N{spStructs} structures; against v3.1 it is better or equal on \N{spFsBeatsTsNewOrgans} (Table~\ref{tab:spineorgans}).
It takes \N{spFsSec}\,s per scan with \N{spFsVRAM}\,GB of GPU memory, against \N{spGpuVRAM}\,GB for the \ts{} network in the same pipeline.
The thoracic and lumbar vertebrae are segmented as well as by \ts; the lower cervical vertebrae (C4--C6) are the weakest structures for all methods, and C6, C1 and the sacrum are where \fs{} is furthest behind \ts{} v3.1.
Keeping only the largest connected component of each vertebra lowers the Dice to \N{spFsLccDice}, so it is not used.

\begin{figure}[t]
\centering
\includegraphics[width=\linewidth]{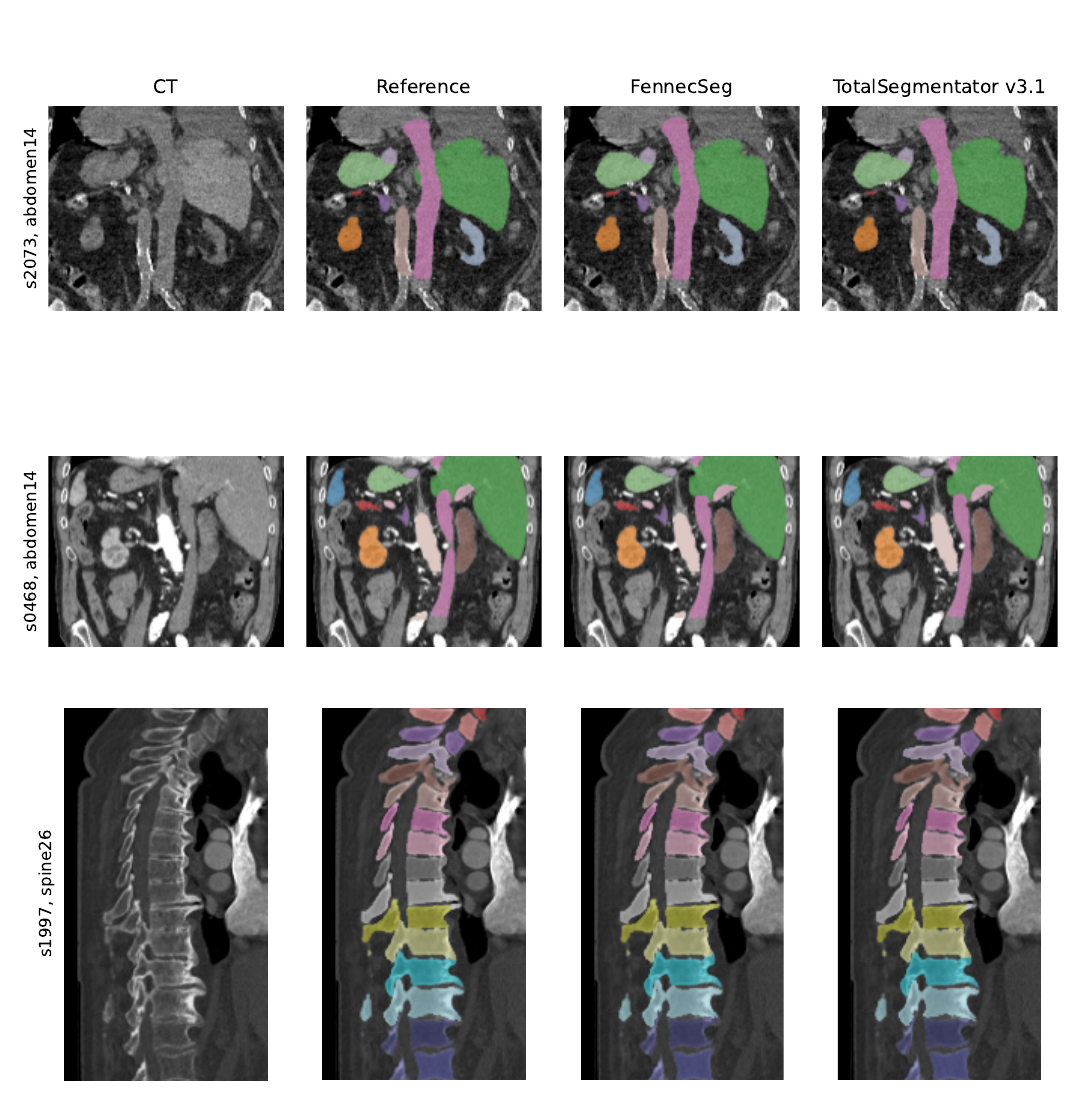}
\caption{Typical test scans (median per-scan Dice of \fs{} among scans showing every structure of the group): reference masks, \fs{} (abdomen14 0.3.0 and spine26 0.1.0, run with the library) and \ts{} v3.1. Coronal slices for the abdomen, a sagittal slice for the spine (thorax not shown); one colour per structure.}
\label{fig:qual}
\end{figure}

\paragraph{Thorax.}
\fs{} thorax7 reaches Dice \N{thFsDice}, \N{thFsGapTsNew} below \ts{} v3.1 and \N{thFsGapGpu} below \ts's v3 organ and cardiac networks in our pipeline, in \N{thFsSec}\,s per scan (\N{xHTsNewOverThFs}$\times$ less than v3.1 per scan, \N{xHTsNewFolderOverThFs}$\times$ less on a folder) with \N{thFsVRAM}\,GB of GPU memory (Table~\ref{tab:thorax}).
It is better or equal on \N{thFsBeatsTsNewOrgans} of the \N{thStructs} structures (left lower lobe, right upper lobe and trachea) and furthest behind on the left upper lobe and the heart, about 0.015 each (Table~\ref{tab:thoraxorgans}).
The right upper lobe is the weakest structure for every method (about 0.82), which suggests that the difficulty lies in the reference masks or the lobe boundary rather than in one model.
No post-processing is applied.
Figure~\ref{fig:qual} shows typical test scans of the abdomen and the spine.

\paragraph{Where the time goes.}
On the first \N{fsScans} test scans, where we also ran the \ts{} 2.18 command in all its modes (Table~\ref{tab:ts218}), the default command takes \N{tsDefaultSec}\,s per scan; the forward passes of its five networks take \N{tsDefaultNetSec}\,s (\N{tsDefaultNetPct}\%) and nnU-Net's pre-processing and export around them \N{tsDefaultCpuPct}\%, the rest being mostly model loading and file input and output, in line with other reports~\citep{nnunet_issue3079,totalsegmentator_v3}.
With the same weights in our pipeline, the Dice is unchanged (\N{gpuAbdDice}) and the two networks for the abdominal structures take \N{gpuAbdSec}\,s, or \N{gpuAllSec}\,s for all five.
\ts{} v3.1 removes most of this overhead in its own code: \N{hTsNewSec}\,s per scan including model loading, and \N{hTsNewFolderSec}\,s on a folder.
The Dice of the 2.18 command and of the same networks in our pipeline is equivalent: over \N{eqScans} scans the mean per-scan difference is \N{eqDiff} (90\% confidence interval $\N{eqLo}$ to $\N{eqHi}$), and a two one-sided test rejects a difference larger than 0.001 ($p=\N{eqP}$).

\begin{table}[t]
\centering
\caption{\ts{} 2.18 as released, on the first \N{fsScans} test scans (14 abdominal structures). On these scans \fs{} takes \N{fsSec}\,s at Dice \N{fsDice}, above \ts's small networks run in the same pipeline (\N{gpuSmallDice}), which it beats on \N{fsBeatsSmallOrgans} of the 14 structures.}
\label{tab:ts218}
\begin{tabular}{lrrrr}
\toprule
Method & s/scan & Dice & NSD & GPU GB \\
\midrule
\ts{} 2.18 default (command) & \N{tsDefaultSec} & \N{tsDefaultDice} & \N{tsDefaultNSD} & \N{tsDefaultVRAM} \\
\ts{} 2.18 fast, 3\,mm (command) & \N{tsFastSec} & \N{tsFastDice} & \N{tsFastNSD} & \N{tsFastVRAM} \\
\ts{} 2.18 small (command) & \N{tsSmallSec} & \N{tsSmallDice} & \N{tsSmallNSD} & \N{tsSmallVRAM} \\
\ts{} default, 2 networks (pipeline) & \N{gpuAbdSec} & \N{gpuAbdDice} & \N{gpuAbdNSD} & \N{gpuAbdVRAM} \\
\ts{} small, 2 networks (pipeline) & \N{gpuSmallSec} & \N{gpuSmallDice} & \N{gpuSmallNSD} & \N{gpuSmallVRAM} \\
\fs{} abdomen14 0.2.0 & \N{fsSec} & \N{fsDice} & \N{fsNSD} & \N{fsVRAM} \\
\bottomrule
\end{tabular}
\end{table}

\begin{table}[t]
\centering
\caption{Dice per abdominal structure (\N{hFsScans} test scans). Best value per row in bold. FS: \fs; filter: Section~\ref{sec:post}.}
\label{tab:organs}
\begin{tabular}{lcccc}
\toprule
Structure & TS v3.1 & TS v3.1 + filter & FS 0.2.0 & FS 0.3.0 \\
\midrule
Spleen & \textbf{0.963} & \textbf{0.963} & 0.946 & 0.960 \\
Kidney R & 0.871 & \textbf{0.905} & 0.845 & 0.894 \\
Kidney L & 0.900 & 0.895 & 0.864 & \textbf{0.901} \\
Gallbladder & 0.810 & \textbf{0.848} & 0.822 & 0.836 \\
Liver & \textbf{0.967} & \textbf{0.967} & 0.944 & 0.954 \\
Stomach & \textbf{0.926} & \textbf{0.926} & 0.918 & 0.919 \\
Pancreas & 0.854 & \textbf{0.861} & 0.838 & 0.854 \\
Adrenal R & \textbf{0.849} & 0.833 & 0.814 & 0.822 \\
Adrenal L & \textbf{0.845} & 0.829 & 0.818 & 0.837 \\
Esophagus & 0.897 & 0.905 & 0.885 & \textbf{0.907} \\
Duodenum & \textbf{0.828} & 0.825 & 0.820 & 0.823 \\
Aorta & 0.910 & \textbf{0.940} & 0.912 & \textbf{0.940} \\
IVC & 0.883 & 0.893 & 0.906 & \textbf{0.909} \\
Portal/splenic vein & 0.831 & \textbf{0.833} & 0.809 & 0.826 \\
\midrule
Mean & 0.881 & 0.887 & 0.867 & 0.884 \\
\bottomrule
\end{tabular}

\end{table}

\begin{table}[t]
\centering
\caption{Dice per spine structure (\N{spFsScans} test scans). Best value per row in bold.}
\label{tab:spineorgans}
\small
\begin{tabular}{lccc}
\toprule
Structure & TS v3.1 & TS v3 net & FennecSeg \\
\midrule
C1 & 0.909 & \textbf{0.913} & 0.874 \\
C2 & 0.841 & \textbf{0.911} & 0.883 \\
C3 & 0.846 & 0.840 & \textbf{0.850} \\
C4 & \textbf{0.783} & 0.781 & 0.754 \\
C5 & \textbf{0.804} & 0.744 & 0.784 \\
C6 & \textbf{0.801} & 0.786 & 0.748 \\
C7 & \textbf{0.901} & 0.860 & 0.866 \\
T1 & \textbf{0.931} & 0.890 & 0.928 \\
T2 & \textbf{0.932} & 0.923 & 0.931 \\
T3 & \textbf{0.909} & 0.898 & 0.903 \\
T4 & 0.887 & \textbf{0.916} & 0.893 \\
T5 & \textbf{0.919} & 0.873 & 0.906 \\
T6 & 0.891 & 0.870 & \textbf{0.907} \\
T7 & 0.892 & 0.891 & \textbf{0.902} \\
T8 & \textbf{0.904} & 0.891 & 0.900 \\
T9 & \textbf{0.935} & 0.914 & \textbf{0.935} \\
T10 & \textbf{0.938} & 0.917 & 0.929 \\
T11 & 0.932 & 0.922 & \textbf{0.946} \\
T12 & 0.888 & 0.916 & \textbf{0.930} \\
L1 & 0.915 & 0.912 & \textbf{0.948} \\
L2 & 0.941 & 0.936 & \textbf{0.954} \\
L3 & \textbf{0.948} & 0.896 & 0.918 \\
L4 & 0.930 & 0.927 & \textbf{0.935} \\
L5 & 0.933 & 0.931 & \textbf{0.936} \\
Sacrum & 0.928 & \textbf{0.941} & 0.893 \\
\midrule
Mean & 0.898 & 0.888 & 0.894 \\
\bottomrule
\end{tabular}

\end{table}

\begin{table}[t]
\centering
\caption{Dice per thorax structure (\N{thFsScans} test scans). Best value per row in bold.}
\label{tab:thoraxorgans}
\small
\begin{tabular}{lccc}
\toprule
Structure & TS v3.1 & TS v3 nets & FennecSeg \\
\midrule
Upper lobe L & \textbf{0.951} & 0.948 & 0.935 \\
Lower lobe L & 0.933 & 0.934 & \textbf{0.942} \\
Upper lobe R & 0.822 & \textbf{0.825} & \textbf{0.825} \\
Middle lobe R & \textbf{0.943} & 0.941 & 0.933 \\
Lower lobe R & \textbf{0.959} & 0.947 & 0.956 \\
Trachea & 0.910 & \textbf{0.929} & 0.911 \\
Heart & 0.927 & \textbf{0.928} & 0.913 \\
\midrule
Mean & 0.921 & 0.922 & 0.917 \\
\bottomrule
\end{tabular}

\end{table}

\section{The \texttt{fennecseg} library}
\label{sec:library}

The library runs \fs{} models with PyTorch only; it does not import \ts{} or nnU-Net.
\begin{itemize}
  \item \textbf{Inputs.} NIfTI, a DICOM series (the largest series in a folder), NRRD or MHA, at any spacing and orientation. The label map is written on the grid and in the orientation of the input.
  \item \textbf{Model packages.} A model is a folder with its weights (\texttt{safetensors}) and a manifest (\texttt{model.json}) that holds the architecture, label table, input spacing, intensity statistics, default window settings, post-processing (such as the size filter of Section~\ref{sec:post}) and its benchmark, measured with the library itself. A new group of structures is a new package, not new code.
  \item \textbf{Versioned releases.} Packages are published on the Hugging Face Hub, one folder per group and an immutable tag per version (released: \texttt{abdomen14-v0.3.0}, \texttt{spine26-v0.1.0} and \texttt{thorax7-v0.1.0}); \texttt{-{}-model abdomen14} loads the latest release and \texttt{abdomen14@0.3.0} an exact one.
  \item \textbf{Interfaces.} A command line (\texttt{fennecseg predict scan.nii.gz -o seg.nii.gz -{}-model abdomen14}, with one file per structure on request) and a Python API (\texttt{fennecseg.load}, \texttt{fennecseg.predict}). Labels can be written with the model's own ids or with \ts's ids, so existing \ts{} workflows keep working.
\end{itemize}
The released packages were benchmarked with the library itself: on all \N{hFsScans} test scans its predictions are identical to those of the benchmark code, voxel for voxel, for abdomen14 0.3.0, spine26 0.1.0 and thorax7 0.1.0.
Table~\ref{tab:lib} summarises the checks of the resampling and DICOM paths on one test scan.

\begin{table}[t]
\centering
\caption{Library checks on one test scan (abdomen14 0.1.0, CPU). ``Limit'': Dice of the 1.5\,mm prediction moved to the same grid, the best a 1.5\,mm model can reach there.}
\label{tab:lib}
\begin{tabular}{llrr}
\toprule
Input & Orientation & Dice & Limit \\
\midrule
NIfTI, 1.5\,mm (benchmark grid) & RAS & \multicolumn{2}{r}{99.999\% voxels as the GPU benchmark} \\
NIfTI, 0.8$\times$0.8$\times$1.5\,mm & LPS & 0.908 & 0.911 \\
NIfTI, 0.8$\times$0.8$\times$2.5\,mm & LPS & 0.871 & 0.880 \\
DICOM series, 0.8$\times$0.8$\times$2.5\,mm & LPS & \multicolumn{2}{r}{100\% voxels as the NIfTI input} \\
\bottomrule
\end{tabular}
\end{table}
On \N{clinScans} test scans resampled to clinical grids, the library takes \N{clinOneSec}\,s per scan at 0.8$\times$0.8$\times$1.0\,mm (at most \N{clinOneMax}\,s) and \N{clinTwoSec}\,s at 0.8$\times$0.8$\times$2.5\,mm, with at most \N{clinOneVRAM}\,GB of GPU memory. The network takes about \N{clinOneNet}\,s of this; the rest is reading, resampling and writing the larger files, which is where the library can still be made faster.

\section{Discussion}
\label{sec:discussion}

\paragraph{Model versus pipeline.}
Running \ts's own networks in our pipeline makes them \N{xTsDefaultOverGpuAll}$\times$ faster than the released 2.18 command (all five networks in both), at the same Dice; this confirms, on a consumer GPU, what recent GPU-resident pipelines report~\citep{fastsegmentator2026,yamagishi2026flashsegmentator,isensee2026streaming}.
\ts{} v3.1 brings most of that gain to its own command line (\N{hTsNewSec}\,s per scan).
Large speed-ups over \ts{} therefore come from inference engineering, not from a model, and they are no longer specific to any one tool.

\paragraph{One general network or one small model per group.}
\ts{} v3.1 segments 117 structures with one 102M-parameter network in \N{hTsNewFolderSec}--\N{hTsNewSec}\,s; a \fs{} model segments one group with \N{fsParamsM}M parameters in about one second.
For a user who needs one group, \fs{} is \N{xHTsNewFolderOverSpFs}--\N{xHTsNewOverThFs}$\times$ faster per scan, uses less GPU memory for the abdomen and thorax (and slightly more for the spine: \N{spFsVRAM} against \N{hTsNewVRAM}\,GB), at about the same accuracy on all three groups.
For a user who needs all 117 structures, one network is likely the better choice: running several \fs{} models in sequence adds their times.
The two designs serve different needs, and we do not claim that \fs{} replaces \ts.

\paragraph{Why small region models.}
A model per group of structures costs only what is asked for, fits in 3--6\,GB of GPU memory, and can be trained for a new group in one night on one consumer GPU when warm-started from an existing \fs{} model (\N{gpuHSpine}--\N{gpuHThorax}\,GPU hours for the spine and thorax); the abdominal model, trained from scratch and refined twice, took \N{gpuHAbdTotal}\,GPU hours in all.
This makes it practical to add structures that general models do not cover, from public or in-house data.

\paragraph{Limitations.}
\begin{itemize}
  \item The \ts{} 2.18 command was timed on a subset of the test split (\N{fsScans} scans for all modes, \N{roiAbdScans} for \texttt{roi\_subset}); its accuracy is that of its networks in our pipeline, measured on all scans.
  \item The size filter removes real structures smaller than \N{filtMm}\,mm$^3$ (\N{filtRealRemoved} of about 1,100 present scan--structure pairs on the test split); users who need such small pieces can switch it off.
  \item All results are on one dataset (the \ts{} test split) and one GPU model; external datasets such as AMOS~\citep{ji2022amos} and VerSe~\citep{sekuboyina2021verse}, and other hardware, are needed to show that accuracy and speed transfer.
  \item The benchmark scans are stored at 1.5\,mm, so the benchmark timings do not include resampling; on clinical grids the library needs \N{clinTwoSec}--\N{clinOneSec}\,s per scan (Section~\ref{sec:library}).
  \item The reference masks of the \ts{} dataset were partly produced with model assistance, so agreement with them is not the same as agreement with expert annotation.
  \item \fs{} covers 14 abdominal structures, the spine, the lung lobes, trachea and heart; the other \ts{} structures are future work.
\end{itemize}

\section{Conclusion}
\label{sec:conclusion}

\fs{} provides small CT segmentation models, one per group of structures, that run in about one second per scan on a consumer GPU, together with a library that runs them on any CT.
The spine and thorax models are within 0.005 Dice of \ts{} v3.1 (Dice \N{spFsDice} against \N{spTsNewDice}, and \N{thFsDice} against \N{thTsNewDice}) and so is the abdominal model (\N{hFsDice} against \N{hTsNewDice}, or \N{hTsNewFilterDice} with our filter), with less time per group of structures.
Further groups of structures will be released as versioned models in the same library and model repository (\url{https://github.com/pycadd/FennecSeg}, \url{https://huggingface.co/amine0110/fennecseg}).

\paragraph{Data and acknowledgements.} All models were trained and evaluated on version 3.0.0 of the \ts{} dataset~\citep{wasserthal2023totalsegmentator}, released under CC BY 4.0; we thank its authors for making it public. \fs{} is not affiliated with the \ts{} authors. The models are research tools, not medical devices.

\bibliographystyle{unsrtnat}
\bibliography{references}

\end{document}